\documentclass{article}

\usepackage[accepted]{icml2024}

\usepackage{url}            %
\usepackage{nicefrac}       %
\usepackage{booktabs}       %
\usepackage{microtype}      %
\usepackage{xcolor}         %
\usepackage{tablefootnote}

\usepackage{amsmath}
\usepackage{amssymb}
\usepackage{amsfonts}
\usepackage{bm}
\usepackage{mathtools}
\usepackage{arydshln}

\usepackage{wrapfig}
\usepackage{float}
\usepackage{graphicx}
\usepackage{subcaption}

\usepackage{enumitem}

\usepackage[acronym]{glossaries}

\newacronym{SII}{SII}{Stability-Informed Initialization}

\newacronym{IVP}{IVP}{Initial Value Problem}
\newacronym{ODE}{ODE}{Ordinary Differential Equation}
\newacronym{DAE}{DAE}{Differential Algebraic Equation}

\newacronym{EF}{EF}{Euler forward}
\newacronym{MP}{MP}{Midpoint}
\newacronym{RK3}{RK3}{3rd-order Runge-Kutta}
\newacronym{RK4}{RK4}{4th-order Runge-Kutta}

\newacronym{NN}{NN}{Neural Network}
\newacronym{CNN}{CNN}{Convolutional Neural Network}
\newacronym{RNN}{RNN}{Recurrent Neural Network}
\newacronym{VAE}{VAE}{Variational Autoencoder}
\newacronym{NODE}{neural ODE}{Neural Ordinary Differential Equation}
\newacronym{RSN}{ResNet}{Residual Network}
\newacronym{MLP}{MLP}{Multilayer Perceptron}
\newacronym{CNF}{CNF}{Continuous Normalizing Flow}
\newacronym{SSM}{SSM}{State Space Model}

\newacronym{ELBO}{ELBO}{Evidence Lower Bound}
\newacronym{VPT}{VPT}{Valid Prediction Time}
\newacronym{MSE}{MSE}{Mean-Squared Error}
\newacronym{CE}{CE}{Cross-Entropy}

\newacronym{RELU}{ReLU}{rectifier linear unit}

\newacronym{WLTP}{WLTP}{Worldwide Harmonized Light Vehicles Test Procedure}
\newcommand{\state}{\bm{x}}
\newcommand{\statepred}{\hat{\bm{x}}}
\newcommand{\dstate}{\dot{\bm{x}}}
\newcommand{\inp}{\bm{u}}
\newcommand{\measurement}{\bm{y}}
\newcommand{\measurementpred}{\hat{\bm{y}}}

\newcommand{\statedim}{d_x}
\newcommand{\inpdim}{d_u}
\newcommand{\measurementdim}{d_y}
\newcommand{\hiddendim}[1]{d_{h_{#1}}}

\newcommand{\expectation}{\mathbb{E}}

\newcommand{\statematrix}{\bm{A}}
\newcommand{\inpmatrix}{\bm{B}}

\newcommand{\stepsize}{h}
\newcommand{\eigen}{\lambda}

\newcommand{\slope}{\kappa}
\newcommand{\ortho}{\Pi}
\newcommand{\shape}{\Lambda}

\newcommand{\inpinit}{\nu}
\newcommand{\inpbound}{\zeta}

\newcommand{\indim}{d_{\text{in},i}}
\newcommand{\W}{\bm{W}}
\newcommand{\bias}{\bm{b}}

\newcommand{\activation}{\sigma}
\newcommand{\numlayers}{n}
\newcommand{\X}{\mathbf{x}}

\newcommand{\R}[1]{\mathbb{R}^{{#1}}}
\newcommand{\transpose}{\text{T}}
\newcommand{\inverse}{{-1}}
\newcommand{\uniform}{\mathcal{U}}
\newcommand{\gaussian}{\mathcal{N}}

\newcommand{\concat}{\oplus}

\newcommand{\eye}{I}

\newcommand{\numsamples}{N}
\newcommand{\teacher}{\mathcal{T}}
\newcommand{\sampletime}{\Delta t}

\newcommand{\latent}{\bm{z}}
\newcommand{\dlatent}{\dot{\bm{z}}}

\newcommand{\latentdim}{d_z}

\newcommand{\predhrz}{N}

\graphicspath{ {./figures/} }

\definecolor{cvprblue}{rgb}{0.21,0.49,0.74}
\definecolor{icmlblue}{rgb}{0.22 0.28,0.57}
\usepackage[pagebackref,
breaklinks,
colorlinks,
citecolor=icmlblue,
urlcolor=icmlblue,
linkcolor=icmlblue]{hyperref}

\usepackage[disable,textsize=tiny]{todonotes}

\icmltitlerunning{Stability-Informed Initialization of Neural Ordinary Differential Equations}

\begin{document}

\twocolumn[
\icmltitle{Stability-Informed Initialization of Neural Ordinary Differential Equations}

\icmlsetsymbol{equal}{*}

\begin{icmlauthorlist}
\icmlauthor{Theodor Westny}{yyy}
\icmlauthor{Arman Mohammadi}{yyy}
\icmlauthor{Daniel Jung}{yyy}
\icmlauthor{Erik Frisk}{yyy}
\end{icmlauthorlist}

\icmlaffiliation{yyy}{Department of Electrical Engineering, Linköping University, Linköping, Sweden}

\icmlcorrespondingauthor{Theodor Westny}{theodor.westny@liu.se}

\icmlkeywords{Machine Learning, ICML}

\vskip 0.3in
]

\printAffiliationsAndNotice{}  %

\begin{abstract}
This paper addresses the training of Neural Ordinary Differential Equations (neural ODEs), and in particular explores the interplay between numerical integration techniques, stability regions, step size, and initialization techniques. 
It is shown how the choice of integration technique implicitly regularizes the learned model, and how the solver's corresponding stability region affects training and prediction performance.
From this analysis, a stability-informed parameter initialization technique is introduced.
The effectiveness of the initialization method is displayed across several learning benchmarks and industrial applications.
\end{abstract}

\section{Introduction}
 Learning dynamic systems from observations has a long history of applications, particularly in time-series analysis and prediction.
This research domain is intrinsically connected to classical system identification and statistical model estimation methods~\cite{ljung1999system}. 
While there are many early studies on using machine learning to adapt parameters of continuous functions from observations~\cite{cohen1983absolute, rico1993continuous, anderson1996comparison, gonzalez1998identification}, using deep architectures to learn from sequential data has largely been dominated by \glspl{RNN}~\cite{lecun2015deep}, with notable exceptions~\cite{oord2016wavenet, bai2018empirical, vaswani2017attention, gu2022efficiently}.
However, with the introduction of \glspl{NODE}~\cite{chen2018neuralode}, the idea of learning continuous functions using \glspl{NN} has attracted the attention of several researchers.
While \glspl{NODE} have been successfully applied in image classification~\cite{gholami2019anode, zhang2019anodev2, zhuang2020adaptive, gusak2020towards}, motivated by their syntactic similarities to that of \glspl{RSN}~\cite{chen2018neuralode, gholami2019anode, zhang2019anodev2, queiruga2020continuous, ott2021resnet}, their strong inductive bias towards sequential modeling makes them attractive for applications that deal with sequential data~\cite{rubanova2019latent, yildiz2019ode2vae, kidger2020neural, huang2020learning, chen2021eventfn, westny2023graph, lipman2023flow, verma2024climode}.

This paper examines the impact of numerical solver choice on the performance and parameterization of models.
We demonstrate that by careful consideration of the stability regions of the dynamic system with those of the selected solver, and by analyzing the eigenvalues from linearizing the models, the integration method and step size can significantly limit the space of learnable parameter values of \glspl{NN} trained via stochastic gradient descent methods.
Based on these observations, a stability-informed initialization technique is proposed for \glspl{NODE}, yielding notable enhancements in training efficiency and prediction accuracy.

\subsection{Contributions}
The primary contributions of this paper are:
\begin{itemize}
	\item An investigation into the effects of the numerical integration method on the parameters of \glspl{NODE} and its connection to solver stability regions.
	\item Proposal of an \gls{SII} technique for \glspl{NODE}.
\end{itemize}

Implementations are made publicly available.\footnote[2]{\fontsize{8pt}{9pt}\selectfont \url{https://github.com/westny/neural-stability}}

\subsection{Related Work}
\label{sec:related_work}
The interest in \glspl{NODE} has grown considerably since they were first introduced in \cite{chen2018neuralode}.
In light of their successful application, numerous research for a more comprehensive understanding of \glspl{NODE}~\cite{yan2019robustness, massaroli2020dissecting, zhang2020approximation, gusak2020towards, zhu2022numerical} has been proposed with further improvements in~\cite{gholami2019anode, finlay2020train, zhuang2020adaptive, xia2021heavy, krishnapriyan2022learning}.
The investigations in this paper target the effects of the numerical integration method and step size on the learned model.
A majority of related works that consider these questions mainly focus on the effect on prediction performance.
However, an important aspect to consider when it comes to the effect of solver selection is not only their discretization error but also their stability regions~\cite{ascher1998computer}, something that could be important when alternating solvers.
Motivated by both the theoretical and practical benefits in a deep learning context, our analysis is mainly focused on fixed-step Runge-Kutta methods.
In fact, the limitations of variable-step solvers when training \glspl{NODE} was noted already in~\cite{chen2018neuralode}, some of which have been the target of later research~\cite{gholami2019anode, zhang2019anodev2, zhuang2020adaptive}.

It was observed in \cite{zhuang2020adaptive, gusak2020towards, mohammadi2023analysis} that exchanging the numerical solver during testing from the one used during training considerably influenced performance.
In \cite{zhuang2020adaptive}, they demonstrated that training a \gls{NODE} with an adaptive step solver and then utilizing fixed-step solvers of first, second, and fourth-order during inference led to a rise in error rates, albeit with a smaller increase as the order increased.
In \cite{queiruga2020continuous} and \cite{ott2021resnet}, these observations were explored in depth.
In \cite{ott2021resnet}, it was observed that employing coarse discretization during training, followed by testing with an equally or less precise solver led to a considerable drop in performance.

Neural ODEs share several similarities with Deep \glspl{SSM}, notably by using a continuous-time formulation.
In the study by \cite{gu2020hippo}, an initialization technique for \glspl{SSM} that enhances their ability to capture long-range dependencies was introduced---effectively integrated into the S4 model \cite{gu2022efficiently}.
However, S4 was observed to suffer from numerical instability during autoregressive generation, which was subsequently addressed in \cite{goel2022sashimi} by ensuring that the state matrix is constrained to be Hurwitz.
Similarly, one can also draw connections between this work and methods in the reservoir computing literature~\cite{lukovsevivcius2009reservoir} as well as the choice of stable dynamical systems in Random Projection Filter Banks~\cite{farahmand2017random}.
While these approaches share a thematic resemblance with our emphasis on stability, a key distinction of our work is the additional focus on the numerical solvers and their stability regions---an essential aspect when solving differential equations numerically and therefore also for the training and prediction processes.

\section{Stability of Dynamic Systems and Solvers}
\label{sec:system_stability}

A continuous-time dynamic model is described in ODE form by a state-transition function $f$ as
\begin{equation}
    \label{eq:neuralODE}
    \dot{\state} = f(\state, \inp; \theta)
\end{equation}
where $\state \in \R{{\statedim}}$ is the dynamic state, and $\inp \in \R{{\inpdim}}$ an external (control) input, and $\theta$ the parameters of the model.
Two important stability properties that are beneficial to consider when learning the function $f$, in particular when $f$ is represented by a deep \gls{NN}, are 
1) the stability of the continuous-time system \eqref{eq:neuralODE}, and 2) the stability region of the numerical solver used to integrate \eqref{eq:neuralODE}.

\subsection{Stability of Dynamic Systems}
Stability in dynamic systems refers to the property of a system to return to its equilibrium state or to maintain its state of equilibrium after being subjected to disturbances~\cite{ljung1999system}.
In the context of control systems, stability is crucial to ensure the desired response and performance.
The stability of a general dynamic system like \eqref{eq:neuralODE} is complex \cite{khalil2020nonlinear}.
However, for our purposes, it will prove sufficient to consider the stability of
\begin{equation}
    \label{eq:linear_state_space}
    \dstate = \statematrix \state + \inpmatrix\inp,
\end{equation}
where the stability is directly related to the eigenvalues $\eigen$ of the matrix $\statematrix$ and exponential stability is ensured if all eigenvalues lie in the complex left half-plane $\eigen < 0$.

\begin{figure}[t]
    \centering
    \includegraphics[width=0.6\columnwidth]{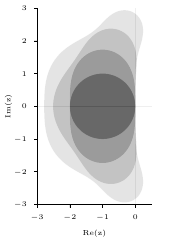}
    \caption{Stability regions for $p\in\{1, 2, 3, 4\}$-stage explicit RK methods of order $p$ where $z=h\lambda$.
    The innermost circle represents the region of stability for the EF method, where $p=1$.
    As $p$ increases, so does the stability region.}
    \label{fig:solver_stability_regions}
\end{figure}

\subsection{Stability Regions of Numerical Solvers}
Differential equations like \eqref{eq:neuralODE} do not, in general, have explicit analytical solutions and numerical integration techniques have to be used \cite{ascher1998computer}. 
Similarly to the stability of the system itself, it is important to consider the stability of the solver as this has a profound impact on the accuracy and stability of the solution trajectory. There are strong relationships between the properties of the system, the chosen solver, the step size, and the accuracy and stability of obtained numerical solutions.

Figure~\ref{fig:solver_stability_regions} illustrates the regions of absolute stability of explicit $p$th-order, $s=p\leq4$ stage Runge-Kutta (RK) methods, where the 1-stage RK method is known as \gls{EF}.
The stability regions are given by the expression
\begin{equation}
    \label{eq:stability_condition}
    \left| 1 + \stepsize \eigen + \frac{(\stepsize \eigen)^2}{2} + \cdots + \frac{(\stepsize \eigen)^p}{p!} \right| \leq 1,
\end{equation}
where $\stepsize$ is the step size and $\eigen$ is the eigenvalue~\citep[pp.~87--89]{ascher1998computer}. From Figure~\ref{fig:solver_stability_regions} it can, e.g., be deduced which maximum step size $\stepsize $ is possible given the dynamics of a system. Here, only fixed-step explicit methods are considered, but the approach can be directly extended to other integration methods. 
For example, several implicit solvers have well-defined stability regions~\citep[p.~143]{ascher1998computer}.

\section{Implications for Neural ODEs}
\label{sec:implications}
\begin{figure}[t]
    \centering
    \includegraphics[width=\columnwidth]{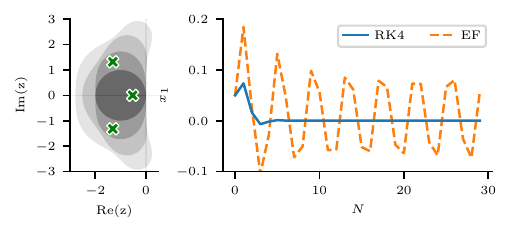}
    \caption{The figure depicts the scaled poles (green crosses) of a synthetic dynamic system $f$ and the simulated response to a small perturbation using a \gls{RK4} and \gls{EF} method.}
    \label{fig:rk4_euler_poles}
\end{figure}

The concept of stability regions is an important property of numerical ODE solvers~\cite{ascher1998computer}.
However, these stability regions and their implications on the learning and prediction performance of \glspl{NODE} have yet to be investigated in detail.

Consider a dynamic system $\dot{\state} = f(\state)$, modeled using a feedforward \gls{NN} with nonlinear activations.
In Figure~\ref{fig:rk4_euler_poles}, we illustrate the poles of the linearized system within the complex plane, i.e., the eigenvalues of the Jacobian matrix $\statematrix = \frac{\partial f(\state)}{\partial \state} |_{\state = \bm{0}}$, scaled by the step size $\stepsize$. 
Clearly, the eigenvalues are well within the stability region of \gls{RK4} and simulating it using that particular solver yields a stable response.
However, for the \gls{EF} method, the two complex eigenvalues are outside of its stability region, and the simulated response exhibits oscillatory behavior resulting in large prediction errors. 
This is an important property to highlight in the context of \glspl{NODE}, since training a model using a higher-order method might yield a model with eigenmodes that are problematic for a lower-order method.
For example, in \cite{queiruga2020continuous, ott2021resnet}  it was observed that changing the numerical solver during inference from the solver employed during training can have a profound impact on the overall performance. 
While they concluded that the behavior was due to the discretization error, this could also be attributed to the solvers' stability regions~\cite{mohammadi2023analysis}.

\subsection{Learned Model Dynamics}
\label{sec:node_solver}
\begin{figure}[!ht]
    \centering
    \includegraphics[width=1\columnwidth]{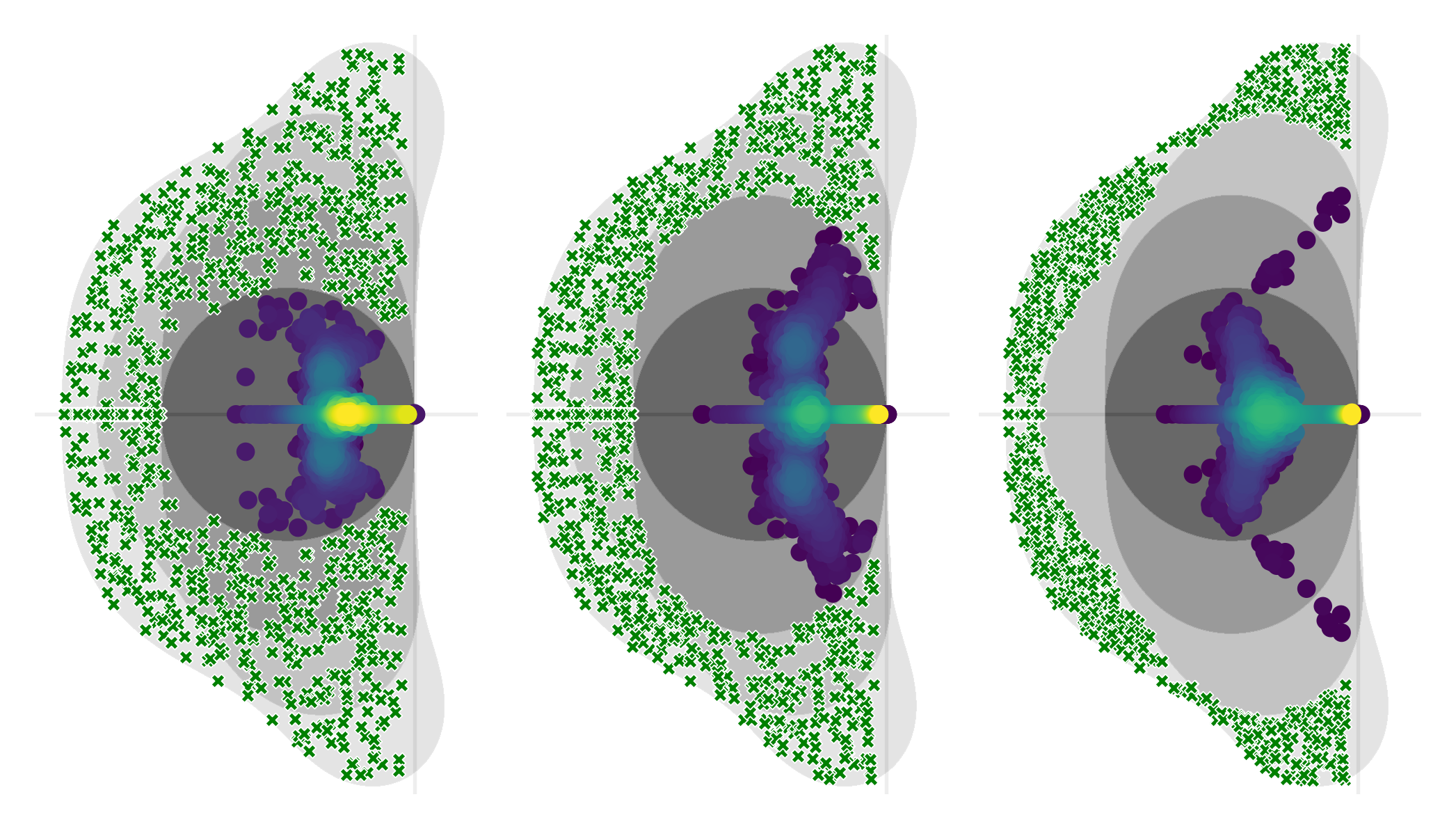}
    \caption{Kernel density estimate of learned model poles based on the approximate linearized system when using the \gls{EF}, \gls{MP}, and an \gls{RK3} method (from left to right).
    The kernel density estimates are based on a total of $3\cdot400$ poles (no. of states $\times$ no. of models).
    The references are various linear systems with $3$ states.
    The combined poles of all linear systems are illustrated with green crosses.}
    \label{fig:kde_methods}
\end{figure}

The choice of numerical solver affects the placement of the learned model dynamics.
To illustrate, Figure~\ref{fig:kde_methods} shows the kernel density estimate of the eigenvalues, for a linearization around a zero reference of $400$ different \glspl{NODE}, depicted in the complex plane.
The models are parametrized using feedforward \glspl{NN} and trained using the \gls{EF}, \gls{MP}, and \gls{RK3} methods, respectively.
They are all tasked with learning the dynamics of a linear system (one unique reference system for each model) with $3$ dynamic states that are unattainable for each solver since the eigenvalues of the state-transition jacobian of the reference systems, shown with green crosses, are manually placed outside of the respective solvers' stability region. With some slight abuse of notation, the eigenvalues of the state-transition jacobian will be referred to as the system poles. 

The training and test data consist of $\numsamples=100$ long sequences, generated by simulating the reference system using \gls{RK4}.
The poles of the learned models used for the illustrations are based on the version of the model that achieves minimum test loss during training.
Several important conclusions can be drawn from the observed results.

Regardless of the nature of the system that the model should learn, the poles are contained within the stability region of the solver.
This follows intuition since poles outside of the respective stability regions would yield poor prediction performance (cf. Section~\ref{sec:implications}).

It is evident from the illustrations and observations made during training that the models display a limited tendency to shift poles to the left within the complex plane.
Note that poles with large negative real parts give rise to states with fast dynamics, and quick transient responses, something the models struggle to learn.
We hypothesize that this phenomenon is related to the spectral bias of \glspl{NN}~\cite{rahaman2019spectral, basri2020frequency, tancik2020fourier}, as it has been observed that \glspl{NN} encounter difficulties when learning higher frequencies. Lack of excitation of high-frequency dynamics could also be a contributing factor \cite{ljung1999system}.

Another important observation to note is the high density of poles around the origin.
The cause of this can be traced back to the initialization technique employed here --- also the default in PyTorch~\cite{paszke2019pytorch}.
In this method, parameter values are initialized from $\uniform(-1/\sqrt{\indim}, 1/\sqrt{\indim})$, where $\indim$ is the input dimension of the $i$-th layer.
When initializing using this method, the linearized system poles are contained within a ball around the origin with a radius dependent on the number of parameters.
This makes learning dynamic systems more difficult since they all have to be moved from the origin during training. Further, with such an initialization strategy some poles lie in the right half-plane, thus rendering the system unstable which makes training more difficult.
An important consideration here is also the effect of using regularization strategies for \glspl{NODE} since penalties on the learnable parameters would have the effect of pushing the poles closer to the origin.

These findings motivate the potential need to develop initialization strategies more suitable for \glspl{NODE} used to learn dynamic systems.
Since the poles of these models are approximately contained within the stability region of the employed solver, it makes intuitive sense to initialize the model within that same region, while also adhering to the stability requirements of dynamic systems.
Furthermore, since the model is stable from the start of training, this allows for training with longer prediction horizons without risking divergence.
In addition, motivated by the increased difficulty of learning fast dynamics, we hypothesize that by initializing the model with poles further into the left-half plane the model will experience a smoother learning process.

\section{Stability-Informed Initialization}
\label{sec:ssi}
Consider a dynamic system in the form \eqref{eq:neuralODE} where $f$ is assumed to be a feedforward \gls{NN} with depth $\numlayers$ in the form

\begin{equation}
    \label{eq:feedforward}
    f(\X) = \W_\numlayers \activation(\cdots \activation(\W_1\X + \bias_1) \cdots) + \bias_\numlayers,
\end{equation}

where $\activation$ is any nonlinearity (activation function), $\W_i\in\R{\hiddendim{i} \times \hiddendim{i-1}}$ and $\bias_i\in\R{\hiddendim{i}}$ are the learnable weights and biases of the model, and $\hiddendim{i}$ is the hidden dimension of layer $i \in \{1,\dots, n\}$. It is assumed that no layer has a dimension less than the state dimension $\statedim$.
Finally, the \gls{NN} input is $\X = \state \concat \inp$, where $\concat$ is the concatenation operation.

Uninformed random initialization of the weights $\W_i$ and biases $\bias_i$ will with high probability give an initial model that violates the stability constraints highlighted in Section~\ref{sec:system_stability}.
As detailed in Section~\ref{sec:implications}, this has a considerable impact on training and will be further discussed in the remainder of the paper.
Now, a rejection sampling-based initialization procedure of the parameters in~\eqref{eq:feedforward} is described that respects the stability region of the chosen solver.

\subsection{Parameter Initialization}
\label{sec:param_init}
The key idea of the approach is to consider a linearization of~\eqref{eq:feedforward} for sufficiently small $\bias_{i}$ and assume an activation function $\activation$ with a linear region.
This assumption holds for several widely used activation functions such as the ReLU and its variants.
For activation functions with variable slopes, like sigmoid or tanh, the linear coefficient can be conservatively replaced by their maximum slope at the origin.
This scales the eigenvalues towards the origin while still maintaining the stability properties of the system.

Note that the stability of the linearized system does not necessarily imply the stability of the nonlinear system, or even of a linear time-varying system, as discussed in~\cite{khalil2020nonlinear}. However, analyzing the stability of the linearized system can provide a useful starting point for initialization.

Let $\slope$ be the slope of the activation function around the origin, then it holds that 
\begin{equation}
    \label{eq:ff_linearization}
    \frac{\partial f(\state)}{\partial \state} \approx \slope^\numlayers\,\W_\numlayers \W_{\numlayers-1} \cdots \W_1 = \statematrix,
\end{equation}
which is exactly an $\numlayers$-layer \emph{linear} feedforward network. This expression is central to the initialization technique---outlined next. Without loss of generality and for simplicity of the presentation, $\slope$ is from now on assumed 1.
Note that, $\W_i$ should be scaled by $1/\slope_i$ if $\slope_i \neq 1$.

\paragraph{Weight matrices}
To illustrate the procedure, first, consider the simple case where the desired eigenvalues of $\statematrix$ in \eqref{eq:ff_linearization} are $\eigen_j\in\mathbb{R}$ and let $\W_i \in \R{\statedim \times \statedim}$ be all identical diagonal square matrices
\begin{equation}
    \label{eq:real_poles}
    \W_{1} = \cdots = \W_\numlayers = \text{diag}(\eigen_1, \eigen_2, \dots, \eigen_{\statedim})^{1/\numlayers}. %
\end{equation}
Then, clearly, $\statematrix$ in \eqref{eq:ff_linearization} has the desired eigenvalues $\eigen_j$.
The $\numlayers$th root is used to allocate information about each eigenvalue throughout the whole feedforward network.
This constitutes one alternative to initializing the weights $\W_i$ but the approach has several restrictions, described next.

First, consider the case when one wants the eigenvalues to be complex.
With real weights $\W_i\in\R{\hiddendim{i} \times \hiddendim{i-1}}$ in~\eqref{eq:feedforward}, this requires that complex eigenvalues come in conjugate pairs.
Consequently, when $\statedim$ is odd, the system must have a least one real eigenvalue. 
Regardless, the construction is still straightforward. Let $\eigen_k\in\mathbb{C}$ and, as before, compute the $\numlayers$th (principal) root as $\eigen_k^{1/\numlayers}= \mu_k + j \omega_k$ where $j$ is the imaginary unit. First, note that the matrix
\begin{equation}
    \label{eq:complex_conjugate_matrix}
    J_{\eigen_k} = 
    \begin{bmatrix}
        \mu_k & \omega_k \\
        -\omega_k & \mu_k 
    \end{bmatrix}
\end{equation}
has exactly the eigenvalues $\mu_k \pm j\omega_k$.
Next, block-diagonal $\W_i$:s are constructed where the $k$th block $J_{\eigen_k}$ from above yields

\begin{align}
\begin{split}
    \label{eq:complex_poles}
    \W_{1} = \cdots = \W_\numlayers = 
    J_{\eigen_1} \concat J_{\eigen_2} \concat \cdots \concat J_{\eigen_{\statedim / 2}} = \\ =\begin{bmatrix}
        \mu_1 & \omega_1 & 0 & \cdots & 0 \\
        -\omega_1 & \mu_1 & 0 & \cdots & 0 \\
        \vdots & \vdots & \vdots & \ddots & \vdots \\
        0 & 0 & 0 & \mu_{\statedim/2} & \omega_{\statedim/2} \\
        0 & 0 & 0 & -\omega_{\statedim/2} & \mu_{\statedim/2} \\ 
    \end{bmatrix}
\end{split}
\end{align}

where $\W_i$ has all real entries and the compound $\statematrix$ has the desired eigenvalues.

Next, we need to address the dimensionality and sparsity of the weight matrices.
Certainly, $\text{dim}(\W_i) \neq \text{dim}(\statematrix)$ is most often the case when constructing a feedforward network like \eqref{eq:feedforward}.
Additionally, to avoid sparse weight matrices where there is a risk of slow learning due to the vanishing gradient problem~\cite{basodi2020gradient, pascanu2013difficulty}, we would like to avoid matrix elements with a value of zero.
Finally, similar to established initialization techniques~\cite{glorot2010understanding, saxe2014exact, he2015delving}, adding randomness to the weight initialization assures that diverse models can be generated, but also crucially breaks symmetry among neurons---enabling diverse feature extraction.
To meet these requirements, the previous ideas presented above are generalized as
\begin{equation}
    \label{eq:ssi}
    \statematrix =  \underbrace{\shape_\numlayers \ortho_{\numlayers-1}^\inverse}_{\W_{\numlayers}} \underbrace{\ortho_{\numlayers-1} \shape_{\numlayers-1} \ortho_{\numlayers-2}^\inverse}_{\W_{\numlayers-1}} \cdots \underbrace{\ortho_2 \shape_2 \ortho_1^\inverse}_{\W_2} \underbrace{\ortho_1 \shape_1}_{\W_1},
\end{equation}
where $\shape_{i}$ is used to address the dimension property and $\ortho_i$ the denseness and randomness property.

The matrices $\shape_i\in\R{\hiddendim{i} \times \hiddendim{i-1}}$ are constructed to inherit the properties of~\eqref{eq:real_poles} and \eqref{eq:complex_poles} for real and complex eigenvalues respectively such that they contain a factor of the system eigenvalues.
To ensure that the dimensions of the respective matrix products adhere to the hidden dimensions of the layers in the network, the matrices are extended with blocks of zeros.
Consider for example the simple case when $\statedim=2$ and $\hiddendim{1}=4$, then
\begin{equation}
    \label{eq:shape_example}
    \shape_1 = 
    \begin{bmatrix}
        \mu_1 & \omega_1 & 0 & 0 \\
        -\omega_1 & \mu_1 & 0 & 0 \\
    \end{bmatrix}^\transpose
\end{equation}
The matrices $\ortho_i \in\R{\hiddendim{i} \times \hiddendim{i}}$ are random orthogonal matrices sampled from the Haar distribution~\cite{stewart1980efficient,mezzadri2006generate}.
In practice, these are constructed by applying QR decomposition of a $\hiddendim{i} \times \hiddendim{i}$ matrix whose elements are independently normally distributed with zero mean and unit variance~\cite{mezzadri2006generate}.
The choice of using orthogonal matrices is natural since their eigenvalues being $1$ ensures that the complete model's eigenvalues are neither attenuated nor amplified along the computational graph and their property of having their transpose equal to their inverse offers numerical and implementational benefits.
Note that introducing the random matrices $\ortho_i$ does not change \eqref{eq:ssi} since
\begin{equation}
    \statematrix =  \shape_\numlayers \ortho_{\numlayers-1}^\inverse \cdots \ortho_2 \shape_2 \ortho_1^\inverse \ortho_1 \shape_1 = \shape_\numlayers  \cdots \shape_2 \shape_1,
\end{equation}
although their inclusion assures that $\forall \text{w}_{j,k} \in \W_i, \text{w}_{j,k} \neq 0$ with probability 1.
Although not explored in this work, we note that it is possible to use other eigenvectors than the standard basis when constructing $\statematrix$.

\paragraph{Dynamic inputs} To include external inputs, the vector $\inp$ is assumed to be concatenated with the state vector $\state$ prior to propagation.
This is easily accounted for by modifying $\shape_1$ such that $\shape_1\in\R{\hiddendim{1} \times (\statedim + \inpdim)}$.
Continuing the example in~\eqref{eq:shape_example}, with $\inpdim=1$, then
\begin{equation}
    \label{eq:shape_inputs}
    \shape_1 = 
    \begin{bmatrix}
        \mu_1 & \omega_1 & 0 & 0 \\
        -\omega_1 & \mu_1 & 0 & 0 \\
        \inpinit_1 & \inpinit_2 & 0 & 0 \\
    \end{bmatrix}^\transpose,
\end{equation}
where $\inpinit_k \sim \uniform(-\inpbound, \inpbound)$ and $\inpbound = \frac{1}{n}\sum_{i = n} |\mu_i|$.

\paragraph{Bias terms} Given that the linearization holds for $\bias_i$ close to $\bm{0}$, it might seem appealing to initialize $\bias_i = \bm{0}$ in order to satisfy the approximation.
However, this could lead to ineffective learning of $\bias_i$.
To that end, initialization of the bias terms $\bias_i$ is done according to
\begin{equation}
    \label{eq:bias_init}
    \bias_i \sim \uniform(-\varepsilon \cdot \bm{1}_{\hiddendim{i}}, \varepsilon \cdot \bm{1}_{\hiddendim{i}}),
\end{equation}
where $\bm{1}_{\hiddendim{i}}$ is a vector of ones with size $\hiddendim{i}$ and $\varepsilon$ is a tunable (small) value, here set to $\varepsilon={10}^{-4}$.

\paragraph{Summary}
Now that a method to randomly initialize the feedforward network~\eqref{eq:feedforward} given a set of eigenvalues has been outlined, the procedure is completed using a simple rejection sampling algorithm.
This involves randomizing real and/or complex-conjugate eigenvalue pairs within the stability region of the selected solver for a given step size based on condition~\eqref{eq:stability_condition}.
This procedure is outlined in Algorithm~\ref{alg:rejection}.
The rejection-sampling algorithm produces a set of eigenvalues that are utilized to form $\shape_i$ as described in \eqref{eq:shape_example} and \eqref{eq:shape_inputs}.
Subsequently, random orthogonal matrices $\ortho_i$ are sampled.
These matrices, together with $\shape_i$ are used to construct the weights $W_i$ as outlined in \eqref{eq:ssi}.
The process concludes with the construction of the bias terms as in \eqref{eq:bias_init}.

\begin{algorithm}[tb]
    \caption{Stability Region Rejection Sampling}
    \label{alg:rejection}
 \begin{algorithmic}
    \STATE {\bfseries Input:} Solver order $p$, step size $\stepsize$, state dimension $\statedim$
    \STATE Initialize: $\mathcal{E} \gets \{\}$
    \REPEAT
    \STATE Sample: $\mu \sim \uniform(-3.0, -0.1)$
    \IF{$|\mathcal{E}| < \statedim - 2$ and \texttt{\small use\_complex}}
        \STATE Sample: $\omega \sim \uniform(-3.0, 3.0)$
    \ELSE
        \STATE Set: $\omega = 0$
    \ENDIF
    \STATE Form: $\eigen = \mu + j\omega$
    \IF{$\left|1 + \eigen + \frac{\eigen^2}{2} + \cdots + \frac{\eigen^p}{p!}\right| < 1 - \epsilon$}
        \STATE $\mathcal{E} \gets \mathcal{E} \cup \{\eigen / h\}$
        \IF{$\omega \neq 0$}
            \STATE $\mathcal{E} \gets \mathcal{E} \cup \{\bar{\eigen} / h\}$
        \ENDIF
    \ELSE 
        \STATE Reject and proceed to the next iteration
    \ENDIF
    \UNTIL{$|\mathcal{E}| = \statedim$}
    \STATE {\bfseries Return:} $\mathcal{E}$
 \end{algorithmic}
 \end{algorithm}

\section{Teacher-Student Regression}
\label{sec:teacher-student}
\paragraph{Model} We examine a teacher-student \gls{NN} setting in which a student network is trained to replicate the output of a teacher network.
The teacher is designed to simulate a nonlinear dynamic system and is adaptable to various dynamic states and inputs, allowing for a rich simulation study.
Both the teacher and student models are outfitted with nonlinear ELU~\cite{clevert2015fast} activation functions. %
The main considered input was in the form of pulse-width modulated waveforms $w$, meant to model a continuous-time input that excites model dynamics.
The training and test set inputs $\inp$ are constructed using different periodic functions with varying frequencies together with added noise
\begin{equation}
    \inp = [w_k + v_k \quad  \dots \quad w_{k+\numsamples} + v_{k+\numsamples}],
\end{equation}
where $v\sim\gaussian(0, 0.1)$ and $\numsamples$ is the number of samples.
Data generation then proceeds by solving the \gls{IVP}
\begin{equation}
    \state = \state_0 + \int_{0}^t\teacher(\tau, \state(\tau), \inp(\tau)) \text{d}\tau, \quad t=\numsamples \cdot \sampletime
\end{equation}
where $\teacher$ is the teacher model, $\state_0 \sim \uniform[0, 1)$ the initial state, and $\state$ the corresponding solution, recovered using a numerical solver, in this case, the variable-step Dormand-Prince method \cite{dormand1980family}.
The state trajectory $\state$ is determined by the integration length $t$, where $\sampletime$ is the sampling period.
When solving the \gls{IVP}, the teacher utilizes linear interpolation to compute intermediate input values.

\paragraph{Training} We conduct a study, using $500$ random seeds (and teachers), training various student models using different hyperparameters, solvers, and initialization techniques. %

\begin{figure}[!t]
    \centering
    \includegraphics[width=\columnwidth]{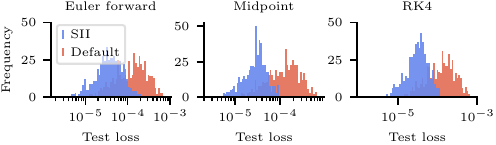}
    \caption{Histogram over minimum test loss when training on $500$ different teachers, initialized with poles within the first-order region.}
    \label{fig:four_state_hist}
\end{figure}

\begin{figure}[t]
    \centering
    \includegraphics[width=\columnwidth]{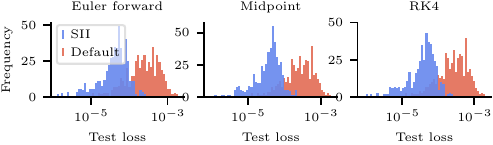}
    \caption{Histogram over minimum test loss when training on $500$ different teachers, initialized with poles outside of all stability regions.}
    \label{fig:two_state_hist}
\end{figure}

\paragraph{Results}
The first investigation is done by initializing the teacher model with $4$ dynamic states using the proposed technique such that the poles are inside of the first-order stability region.
Each student model has the same hyperparameter configuration as the teachers, in this case, $3$ layers with $128$ nodes in each layer.
Despite the randomness in the example, there is a clear tendency toward lower loss values using the proposed technique, with orders of magnitude smaller losses (see Figure~\ref{fig:four_state_hist}).

In general, the eigenvalues of the true system to be learned from training data are rarely known.
To simulate the case when neither the solver nor the step size is chosen optimally, the second investigation is conducted to study how well the nonlinear students can approximate the data regardless.
This is done by initializing the teacher model with $2$ dynamic states using the proposed technique such that the poles are outside of the stability region of the employed solver.
To add to the difficulty of the task, each student model is under-parameterized relative to the teacher.
While the teacher has $3$ layers with $128$ nodes in each layer, the student has $2$ layers with $32$ nodes in each layer.

In Figure~\ref{fig:two_state_hist}, the histograms of the minimum test loss are shown for the different solvers and initializations.
Similar to the first investigation, the results show a clear improvement. %

\section{Experiments}
To assess the performance and usability of the proposed technique, several experiments are conducted across a diverse set of problem domains.
Our choice of experimental cases was guided by three principal criteria: (i) prevalence in the literature, (ii) complexity and diversity of the tasks, and (iii) real-world applicability and relevance.
For each experimental scenario, multiple instances of the models are trained under different random seeds to ensure the reproducibility and statistical significance of the results.
Importantly, the configuration of hyperparameters, e.g., batch size and learning rate may differ with the task but is always the same for all methods. 
Unless stated otherwise, the models are trained and tested using the Midpoint method.

\subsection{Pixel-Level Image Classification}
\label{sec:im-class}
Drawing inspiration from \cite{le2015simple, gu2022efficiently}, the original MNIST~\cite{lecun1998gradient} and CIFAR-10~\cite{krizhevsky2009learning} datasets are modified for a sequential classification task.
This modification involves presenting images to the model one pixel at a time, requiring the model to classify the image only after all pixels have been observed.
These tasks have been popularized as benchmarks for long-range dependency problems.

\paragraph{Model}
We utilize an encoder-decoder-like architecture, embedding each pixel $\inp_k \in [0, 1]$ into a high-dimensional vector $\tilde{\inp}_k \in \R{d_{\tilde{u}}}$ through a linear-affine transformation. %
These vectors are sequentially fed into a \gls{NODE}, which are used to compute the time derivative of the latent state $\latent_k \in \mathbb{R}^{\latentdim}$, expressed as $\dot{\latent} = f(\latent, \tilde{\inp})$, where $f$ is a neural network.
Through numerical integration, the latent state is updated for each new pixel.
After processing all pixels, the sequence's final state is input into a two-layer feedforward network to predict the class probabilities $\measurementpred_N = h(\latent_N)$, where $N$ is the number of pixels.

\paragraph{Training and testing}
We incorporate principles from early-exit networks~\cite{anon2016branchynet} to enhance feature propagation during the encoding stage by training the model to predict the correct class at all intermediate steps.
The loss function is formulated as
\begin{align}
    \label{eq:ce}
    \mathcal{L}_{\rm CE} = \sum_{k=1}^{N} \ell_{CE}\bigl(\measurementpred_{k}, \measurement\bigr) \cdot \bm{w}_k,
\end{align}
where $\ell_{CE}$ is the standard \gls{CE} criterion, and $\bm{w}_k$ is an exponentially increasing term that assigns more importance to later predictions. 
During testing, the model output is taken as the final prediction $\measurementpred_{N}$ in the sequence.

\begin{figure}[t]
    \centering
    \includegraphics[width=\columnwidth]{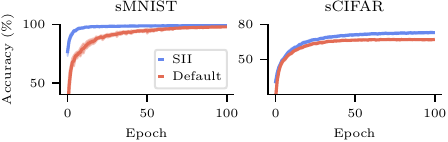}
    \caption{Test accuracy over training epochs on sequential MNIST and CIFAR-10. Solid lines represent the mean test accuracy, and the shaded regions illustrate the approximate $99$\% confidence interval around the mean.}
    \label{fig:smnist}
\end{figure}

\paragraph{Results} For each dataset, $10$ model instances for each initializing technique were trained.
The resulting accuracy is summarized in Table~\ref{tab:seq-im-res}, reporting the mean and standard deviation.
Here, pMNIST refers to the permuted MNIST variant~\cite{goodfellow2013empirical}.
It is clear from the results that using the proposed initializing technique provides positive improvements in test accuracy.
Additionally, employing \gls{SII} shows clear enhancement in training efficiency, achieving convergence at greater speed.
This effect is depicted in Figure~\ref{fig:smnist}, where the test acccuracy curves are illustrated across the training epochs.

\begin{table}[!h]
	\caption{Pixel-Level Classification Accuracy (mean $\pm$ std.)}
	\label{tab:seq-im-res}
	\centering
	\resizebox{\columnwidth}{!}{%
	\begin{tabular}{l l l l}
	  \toprule
	   & sMNIST & spMNIST & sCIFAR \\
	  \midrule
	  Default & $97.9 \pm 0.20$ \% & $90.4 \pm 0.41$ \% & $67.6 \pm 0.15$ \% \\
	  SII & $\bm{99.0} \pm 0.09$ \% & $\bm{94.4} \pm 0.26$ \% & $\bm{73.3} \pm 0.32$ \% \\
	  \bottomrule
	\end{tabular}}
\end{table}

\subsection{Latent Dynamics}
\label{sec:latent-dynamics}
This evaluation centers on benchmarks introduced in \cite{botev2021priors}.
These benchmarks encompass three core tasks: (1) embed a sequence of images ($\state_0, \dots, \state_t$) to a lower-dimensional abstract state $\latent_t$, (2) simulating the system in the latent space using a continuous dynamic model, and (3) mapping the predicted sequence ($\latent_{t+1}, \dots, \latent_{t + \predhrz}$) back into the image domain ($\state_{t+1}, \dots, \state_{t + \predhrz}$).

\paragraph{Model}
While our approach is largely inspired by the proposal in \cite{toth2019hamiltonian}, the model structure used is based on the U-Net architecture~\cite{ronneberger2015unet}.
Given a sequence of images $[\state_0, \dots, \state_t] \in \R{t \times 3 \times 32 \times 32}$, the first step is to transpose the channel and temporal dimension. %
By treating the temporal dimension as an additional feature alongside the image height and width, 3D convolution and pooling operations are utilized for input embedding.
The encoder (contracting path), compresses the image sequence into the abstract state $\latent_t \in \R{\latentdim \times 4 \times 4}$ while simultaneously increasing the channel dimension ($\latentdim \gg 3$).
After forward integration of the initial abstract state, the decoder (expansive path) reconstructs the predicted states into a sequence of images $[\statepred_{t+1}, \dots, \statepred_{t + \predhrz}] \in \R{\predhrz \times 3 \times 32 \times 32}$. 

\paragraph{Training and testing} For this study, a simple summed \gls{MSE} loss is used for training
\begin{align}
    \label{eq:mse}
    \mathcal{L}_{\rm MSE} &= \sum_{k=1}^{\predhrz} \bigl(\statepred_{t + k} - \state_{t + k}\bigr)^2
\end{align}
where $\statepred$ is the prediction and $\state$ is the ground truth image.
During training, we set $\numsamples = 60$ and $t = 10$. 

During testing, the prediction horizon is increased in order to assess the long-term prediction and extrapolation capabilities of the respective models.
For a more fair comparison across datasets, performance is reported using the normalized MSE~\cite{zhong2021benchmarking}:
\begin{equation}
    \text{MSE}_n = \frac{1}{\predhrz}\sum_{k=1}^{\predhrz} \frac{\bigl(\statepred_{t + k} - \state_{t + k}\bigr)^2}{\state_{t + k}^2}
\end{equation}

\paragraph{Results}
Investigations are conducted on four out of the 17 datasets in \cite{botev2021priors}: the Spring-mass system (Mass Spring), the Double Pendulum, Molecular Dynamics (16 particles), and the 3D Room (consisting of MuJoCo~\cite{todorov2012mujoco} scenes).
Ten model instances were trained for each dataset under each initialization technique.
The resulting normalized \gls{MSE} is summarized in Table~\ref{tab:latent}, illustrating clear improvements across all datasets when employing \gls{SII}.
Notably, the experiments revealed that the benefits of \gls{SII} became increasingly evident as the prediction horizon was extended, a trend that can be hypothesized to stem from its stability properties.

\begin{table}[!ht]
	\caption{Latent Dynamics $\text{MSE}_n$ (mean $\pm$ std.)}
	\label{tab:latent}
	\centering
	\begin{tabular}{l c c}
	  \toprule
	   & Default & SII\\
	  \midrule
 	  Mass Spring & $0.102 \pm 0.015$ & $\bm{0.066} \pm 0.008$  \\
	  Double Pend. & $0.084 \pm 0.017$  & $\bm{0.061} \pm 0.002$   \\
	  Molecular Dyn. & $0.151 \pm 0.036$ & $\bm{0.106} \pm 0.011$ \\
	  3D Room & $0.588 \pm 0.102$  & $\bm{0.176} \pm 0.055$ \\
	  \bottomrule
	\end{tabular}%
\end{table}

\subsection{Multivariate Time-Series Forecasting}
\label{sec:case-study}
This study aims to develop a model capable of simulating the dynamics of an unknown system.
The model is tasked with forecasting specific sensor reference values based on signals received from other components within the system:
\begin{subequations}
\begin{align}
    \dlatent &= f_{\gamma}(\latent, \inp) \label{eq:latent_ode_f} \\ 
    \measurementpred &= h_{\theta}(\latent, \inp) \label{eq:latent_ode_h}
\end{align}
\end{subequations}
where $\measurementpred\in\R{\measurementdim}$ is the measurement prediction, and $\inp\in\R{\inpdim}$ is a vector of input measurement signals.
Importantly, the system is partly described by a state-transition function (\ref{eq:latent_ode_f}), with unknown states and dynamics.
In order to compute predictions according to (\ref{eq:latent_ode_h}), the latent states $\latent\in\R{\latentdim}$ must be solved for, which requires that $\latent_0$ is known. 
Inspired by \glspl{VAE}~\cite{diederik2014auto, rezende2014stochastic}, we employ an encoder network $g_{\phi}$ that outputs the variational posterior over the initial state $\latent_0 \sim q_{\phi}(\latent_0 | \inp_0)$ and parametrize it as a diagonal Gaussian with learnable parameters.

The study utilizes three distinct datasets.
The first dataset features standard signals (commonly found in commercial vehicles) from an internal combustion engine, recorded over multiple driving cycles~\cite{jung2022automated}.
The second dataset, referred to as the Human Activity dataset~\cite{vidulin2010localization}, features data from five individuals, each wearing four localization tags (left ankle, right ankle, belt, chest).
The third dataset, referenced as the Air Quality dataset~\cite{saverio2016air} encompasses a set of air quality measurements from an urban monitoring station in Italy.
It provides measurements of various air pollutants and particulate matter, along with environmental factors such as temperature and humidity. 

\begin{figure}[!t]
    \centering
    \includegraphics[width=\columnwidth]{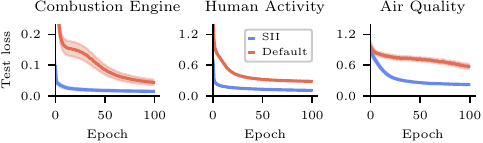}
    \caption{Test loss over the total number of training epochs on the time-series forecasting datasets.
    Solid lines represent the mean test loss and the shaded regions illustrate the approximate $99$\% confidence interval around the mean.}
    \label{fig:case_study_convergence}
\end{figure}

\paragraph{Training and testing} Training targets the optimization of the sequential \gls{ELBO}:
\begin{align}
    \begin{split}
    \label{eq:elbo}
    \mathcal{L}_{\rm ELBO} &= \expectation_{q_\phi(\latent_0 | \inp_0)} \left[\textstyle\sum_{k=1}^{\numsamples} \log p_\theta(\measurement_k | \latent_k, \inp_k) \right] \\ 
    &- D_{\rm KL}\bigl(q_\phi(\latent_0 | \inp_0) \,||\ p(\latent_0)\bigr),
    \end{split}
\end{align}
where $\measurement$ is the ground truth measurement and $p(\latent_0) = \gaussian(\bm{0}, \eye)$ is the prior over the initial latent states.
The omission of $\frac{1}{\numsamples}$ in front of the reconstruction term is done intentionally to force the model to focus more on the prediction error, something that was found to improve validation performance.
Test performance is reported using \gls{MSE}.

\paragraph{Results}
A total of $3 \times 10$ different models per initialization method were trained, consisting of $3$ random data splits (to mitigate some selection bias~\cite{cawley2010over}) with $10$ different model configurations each.
Test performance is presented in Table~\ref{tab:case-study-res}.
Using \gls{SII} results in a significant improvement in test performance compared to the default initialization.
This enhancement is further supported by the faster convergence rate, as depicted in Figure~\ref{fig:case_study_convergence}.

\begin{table}[!h]
	\caption{Multivariate Time-Series \gls{MSE} (mean $\pm$ std.)}
	\label{tab:case-study-res}
	\centering
	\resizebox{\columnwidth}{!}{%
	\begin{tabular}{l c c c}
	  \toprule
	   & Combustion Engine & Human Activity & Air Quality \\
	  \midrule
		Default & $0.04 \pm 0.019$  & $0.29 \pm 0.006$ & $0.57 \pm 0.103$  \\
	  	SII & $\bm{0.01} \pm 0.005$ & $\bm{0.11} \pm 0.003$ & $\bm{0.22} \pm 0.020$  \\
	  \bottomrule
	\end{tabular}}
\end{table}

\section{Scope and limitations}
\label{sec:limitations}

\paragraph{Learnable model} While the proposed technique is specific to a single class of feedforward \glspl{NN}, the method could potentially be extended to any model class with linear operations, e.g., convolutional neural networks.
Furthermore, the hidden dimension of each layer must be equal to or larger than the number of dynamic states to preserve the eigenvalues.

\section{Conclusions}
This paper analyses how the stability properties of the system and stability regions of the chosen numerical solver affect the training and prediction performance of \glspl{NODE}.
Further, it is illustrated how standard techniques for the initialization of the network, without considering the stability properties of the system and solver, may lead to slow training and suboptimal performance.
Based on this, a general stability-informed initialization (\gls{SII}) method is developed that adapts to the stability region of a chosen solver.
The effectiveness of the approach is demonstrated in various machine-learning tasks, including successful applications to real-world measurement data.
In all cases, increased efficiency of training, less susceptibility to random fluctuations, and improved model performance are demonstrated.

\section*{Acknowledgements}
This research was supported by the Strategic Research Area at Linköping-Lund
in Information Technology (ELLIIT) and the Wallenberg AI, Autonomous Systems and Software Program (WASP) funded by the Knut and Alice Wallenberg Foundation.
Computations were enabled by the Berzelius resource provided by the Knut and Alice Wallenberg Foundation at the National Supercomputer Centre.
The authors would like to thank the reviewers for their insightful comments and suggestions, which have significantly improved the manuscript.

\section*{Impact Statement}
This paper presents work whose goal is to advance the field of 
Machine Learning. There are many potential societal consequences 
of our work, none which we feel must be specifically highlighted here.

\bibliography{references}
\bibliographystyle{icml2024}

\end{document}